\pdfoutput=1

\documentclass[11pt]{article}

\usepackage[]{acl}

\usepackage{times}
\usepackage{latexsym}
\usepackage[T1]{fontenc}
\usepackage[utf8]{inputenc}
\usepackage{microtype}
\usepackage{inconsolata}
\usepackage{booktabs}
\usepackage{multirow}
\usepackage{graphicx}
\usepackage{caption}
\usepackage{subcaption}
\usepackage{tabularx}
\usepackage{colortbl}
\usepackage{arydshln}

\title{What Makes Math Word Problems Challenging for LLMs?} 

\author{KV Aditya Srivatsa \\
  MBZUAI \\
  Abu Dhabi, UAE \\
  \texttt{vaibhav.kuchibhotla@mbzuai.ac.ae} \\\And
  Ekaterina Kochmar \\
  MBZUAI \\
  Abu Dhabi, UAE \\
  \texttt{ekaterina.kochmar@mbzuai.ac.ae} \\}

\begin{document}
\maketitle
\begin{abstract}

This paper investigates the question of what makes math word problems (MWPs) in English challenging for large language models (LLMs). We conduct an in-depth analysis of the key linguistic and mathematical characteristics of MWPs. In addition, we train feature-based classifiers to better understand the impact of each feature on the overall difficulty of MWPs for prominent LLMs and investigate whether this helps predict how well LLMs fare against specific categories of MWPs.\footnote{Our code, data, and analysis are publicly available at \href{https://github.com/kvadityasrivatsa/analyzing-llms-for-mwps}{github.com/kvadityasrivatsa/analyzing-llms-for-mwps}}

\end{abstract}

\section{Introduction}

In recent years, large language models (LLMs) have not only demonstrated huge potential across a range of core NLP tasks~\cite[inter alia]{zhao2023survey,brown2020language,radford2019language}, but also exhibited a number of emergent abilities, such as an ability to solve mathematical puzzles~\cite{wei2022emergent}. Math word problems (MWPs) have been proposed as a challenging testbed for LLMs, as they test not only the ability of the models to deal with purely mathematical expressions, but also their reasoning and natural language understanding abilities~\cite[inter alia]{wang2023msat,cobbe2021training,patel-etal-2021-nlp,miao-etal-2020-diverse}. Experiments show that even quite powerful LLMs are still challenged by MWPs~\cite{cobbe2021training}. At the same time, most previous work has either focused on evaluation of LLMs' performance on MWPs or on changes in their behavior in response to progressive-hint prompting, prompt paraphrasing or similar approaches~\cite{norberg2023rewriting,raiyan-etal-2023-math,zheng2023progressive,zhu2023promptbench}, while an in-depth analysis of what exactly makes math problems challenging for LLMs is lacking. We aim to address this knowledge gap.

\begin{figure}[t]
     \centering
     \includegraphics[width=0.95\columnwidth]{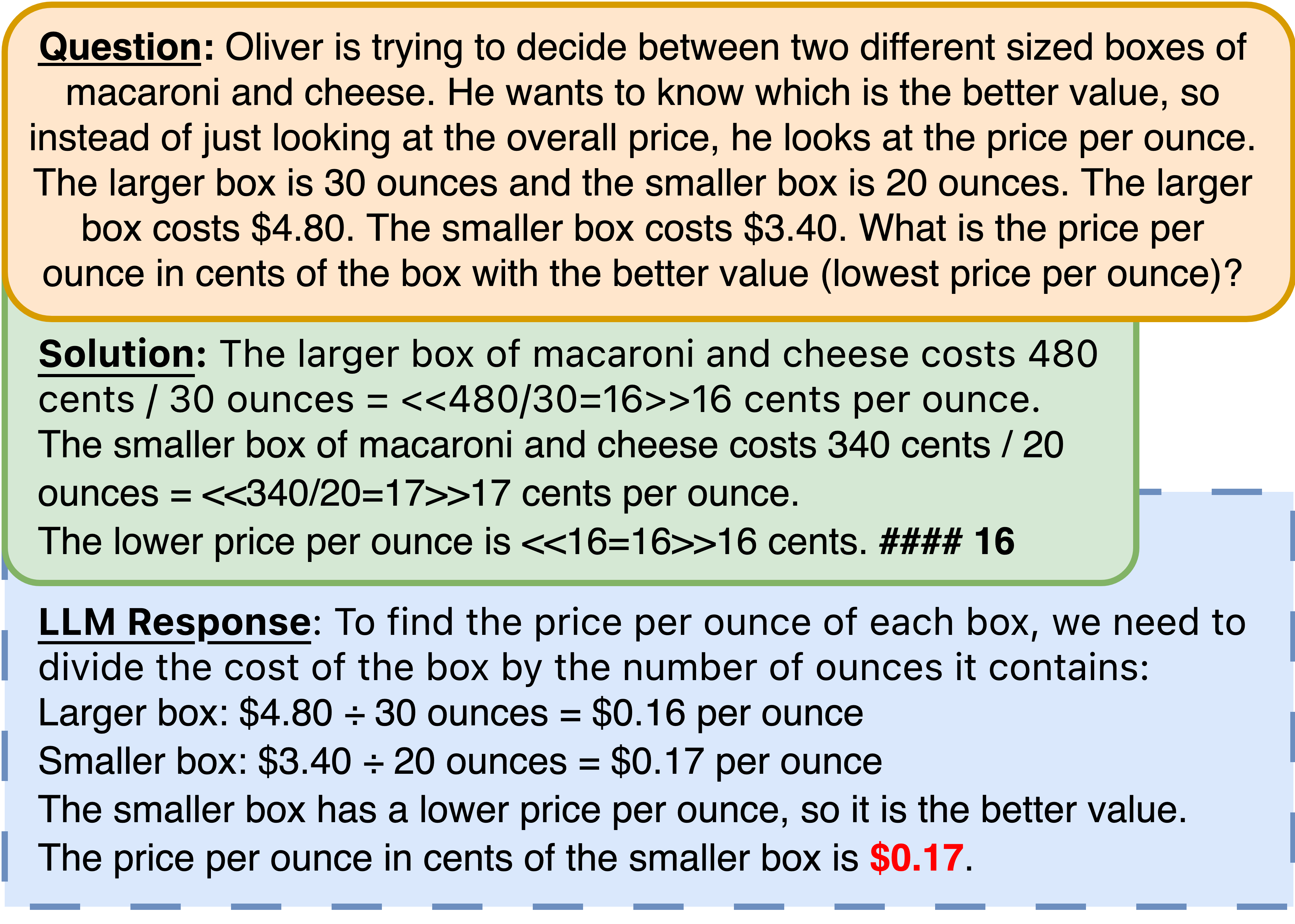}
     \caption{A response from {\tt Llama2-70B} to a lengthy math problem that involves NLU challenges.}
     \label{fig:sample_question}
\end{figure}

A recent study by~\citet{almoubayyed2023instruction} demonstrates a strong connection between reading skills and math outcomes in students. We hypothesize that LLMs' ability to solve MWPs correctly may similarly rely on: (1) {\em the linguistic complexity} of the questions; (2) {\em the conceptual complexity} of the tasks (e.g., the number of steps and types of math operations involved); and (3) {\em the amount of real-world knowledge} required to solve the tasks. Supporting this intuition, our preliminary analysis of the {\tt GSM8K} dataset~\cite{cobbe2021training} suggests that relatively short questions with a small number of described entities, a few calculation steps and a limited range of operators involved in the solution (e.g., {\em Mark is 7 years older than Amy, who is 15. How old will Mark be in 5 years?}) are typically answered correctly by a range of LLMs. At the same time, long questions requiring real-world knowledge (e.g., how many cents there are in a dollar) and extended natural language understanding (NLU) (e.g., interpretation of a lower price) pose challenges for LLMs (see Figure \ref{fig:sample_question}). 

In this paper, we formulate and investigate two research questions: (1) {\em Which characteristics of the input math word questions make them complex for an LLM?} and (2) {\em Based on these characteristics, can we predict whether a particular LLM will be able to solve specific input MWPs correctly?}

\section{Methodology}

\paragraph{Data} We use the \texttt{GSM8K} dataset ~\cite{cobbe2021training}, divided into $7,473$ training and $1,319$ test instances, because of the high quality of human-generated MWPs. This dataset contains a diverse set of problems in English with minimal amount of recurring templates. Furthermore, the difficulty level of the problems is tailored for LLMs, allowing for a wide variation in correctness across models and question types, which is ideal for our feature-based analysis.

\paragraph{Approach} We collect solution attempts from several LLMs to the questions from the {\tt GSM8K} training and test sets. Next, we train statistical classifiers on a filtered subset of questions to predict if they are consistently solved correctly or incorrectly across multiple runs of the models. Our approach is relatively simple 
but it allows us to investigate which of the features are most indicative of the challenges LLMs face in solving math problems.

\paragraph{LLMs} We select an array of open-source models for our experiments. We use {\tt Llama2} (13B and 70B)~\cite{touvron2023llama}, {\tt Mistral-7B} ~\cite{jiang2023mistral} as its performance on math tasks has been found to match models several times its size, and {\tt MetaMath-13B}~\cite{yu2023metamath} as it is fine-tuned on math QA data in contrast to the other general-purpose models in the pool.

\paragraph{Features} We analyze and experiment with the features extracted from MWP questions and their respective expected solutions. This way, the features remain grounded in the dataset, allowing our approach to be applied to any LLM. The features are broadly grouped into the following categories:\footnote{The complete list of features extracted, their description and further statistics can be found in Appendix \ref{sec:appendix_feature_details}} 

\vspace{-0.5em}
\begin{enumerate}
    \item \textbf{Linguistic features} focus on the phrasing of the question. These include the length of the question, sophistication of the vocabulary, syntactic complexity, instances of coreference, and overall readability. Note that the linguistic features are only extracted from the question body as the phrasing of the gold solution has no impact on the expected answer. 
    \vspace{-0.5em}
    \item \textbf{Mathematical features} cover the math arguments, operations, and reasoning steps required to solve the questions. These include the number and diversity of the math operations in the solution body. Arguments provided in the question but not utilized in the solution also require mathematical reasoning for them to be disregarded as noise. Note that while a question can be phrased in many ways (affecting its linguistic features), the underlying math operations and reasoning steps (thus, the mathematical features) remain unchanged.
    \vspace{-0.5em}
    \item \textbf{Real-world knowledge \& NLU based features} indicate the amount of extraneous information needed to solve the task that is not provided explicitly in the question. This may include how many days there are in a month or the interpretation of ``half'' as $1/2$. 
\end{enumerate}

\section{Experiments}

\subsection{Solution Generation} 

To collect solution attempts from the LLMs, we use a simple task-specific prompt (See Appendix \ref{sec:appendix_querying_details}) to minimize any bias imposed on the model generation. We query each LLM $5$ times on each question with varying generation seeds and a temperature of $0.8$. A soft-matching strategy is then used to extract the final answer from the solutions. Using each LLM's attempted solutions, every question is assigned a mean \textbf{success rate} using \texttt{(\# of correct answers) / (\# of solution attempts)}.

\begin{table}[t]
\centering
\resizebox{0.65\columnwidth}{!}{%
\begin{tabular}{@{}lcc@{}}
\toprule
\multicolumn{1}{c}{\multirow{2}{*}{\textbf{Model}}} & \multicolumn{2}{c}{\textbf{Success Rate} ($N$=1,319
)} \\ \cmidrule(l){2-3} 
\multicolumn{1}{c}{} & \textbf{$\mu$} & \textbf{$\sigma$} \\ \midrule
Llama2-13B & 0.3724 & 0.3681 \\ \midrule
Llama2-70B & 0.5609 & 0.3941 \\ \midrule
Mistral-7B & 0.3627 & 0.3309 \\ \midrule
MetaMath-13B & 0.6373 & 0.3816 \\ \bottomrule
\end{tabular}%
}
\caption{Success rates for solution attempts per LLM}
\label{tab:success_rates}
\end{table}

\subsection{Success Rate Prediction} 
We train and evaluate classifiers on their ability to predict for input test questions whether they will be answered correctly or incorrectly by a specific LLM. We also train and evaluate classifiers on the \textbf{intersection} set of questions, which are either solved correctly by {\em all} or by {\em none} of the LLMs. 

\paragraph{Models} We use Logistic Regression, Decision Tree, and Random Forest classifiers, which allow us to extract relative feature importance with ease.
    
\paragraph{Data} For high confidence samples, we use the training and test subset from {\tt GSM8K} where the sampled success rate is either $1.0$ (\textbf{always} correct) or $0.0$ (\textbf{never} correct). The distribution of the LLM-specific splits is detailed in Table 
\ref{tab:class_distr_and_classification}.
    
\paragraph{Preprocessing \& Optimization} We employ several preprocessing steps including dropping highly correlated features, class-balancing, and feature scaling. We also perform a hyperparameter search for each model to maximize performance on unseen data. See Appendix \ref{sec:appendix_training_details} for more details.

\section{Results}

\begin{table*}[]
\centering
\resizebox{0.85\textwidth}{!}{%
\begin{tabular}{ccclclclclcl}
\hline
\multicolumn{12}{l}{} \\
\multicolumn{12}{c}{\textbf{Class Distribution}} \\ \hline
\textbf{Split} & \textbf{Class} & \multicolumn{2}{c}{\textbf{Llama2-13b}} & \multicolumn{2}{c}{\textbf{Llama2-70b}} & \multicolumn{2}{c}{\textbf{Mistral-7b}} & \multicolumn{2}{c}{\textbf{MetaMath-13b}} & \multicolumn{2}{c}{\textbf{Intersection}} \\ \hline
\multirow{3}{*}{Train} & Always & \multicolumn{2}{c}{1102 (30.22\%)} & \multicolumn{2}{c}{2438 (61.36\%)} & \multicolumn{2}{c}{733 (24.06\%)} & \multicolumn{2}{c}{5162 (94.7\%)} & \multicolumn{2}{c}{205 (53.38\%)} \\ \cline{2-12} 
 & Never & \multicolumn{2}{c}{2545 (69.78\%)} & \multicolumn{2}{c}{1535 (38.64\%)} & \multicolumn{2}{c}{2313 (75.94\%)} & \multicolumn{2}{c}{289 (5.3\%)} & \multicolumn{2}{c}{179 (46.61\%)} \\ \cline{2-12} 
 & Total & \multicolumn{2}{c}{3647} & \multicolumn{2}{c}{3973} & \multicolumn{2}{c}{3046} & \multicolumn{2}{c}{5451} & \multicolumn{2}{c}{401} \\ \hline
\multirow{3}{*}{Test} & Always & \multicolumn{2}{c}{188 (28.14\%)} & \multicolumn{2}{c}{427 (60.06\%)} & \multicolumn{2}{c}{111 (21.51\%)} & \multicolumn{2}{c}{528 (71.64\%)} & \multicolumn{2}{c}{31 (24.41\%)} \\ \cline{2-12} 
 & Never & \multicolumn{2}{c}{480 (71.86\%)} & \multicolumn{2}{c}{284 (39.94\%)} & \multicolumn{2}{c}{405 (78.49\%)} & \multicolumn{2}{c}{209 (28.36\%)} & \multicolumn{2}{c}{96 (75.59\%)} \\ \cline{2-12} 
 & Total & \multicolumn{2}{c}{668} & \multicolumn{2}{c}{711} & \multicolumn{2}{c}{516} & \multicolumn{2}{c}{737} & \multicolumn{2}{c}{135} \\ \hline
\multicolumn{12}{l}{} \\
\multicolumn{12}{c}{\textbf{Classification Performance}} \\ \hline
\multicolumn{2}{l}{\multirow{2}{*}{\textbf{Classification Model}}} & \multicolumn{2}{c}{\textbf{Llama2-13b}} & \multicolumn{2}{c}{\textbf{Llama2-70b}} & \multicolumn{2}{c}{\textbf{Mistral-7b}} & \multicolumn{2}{c}{\textbf{MetaMath-13b}} & \multicolumn{2}{c}{\textbf{Intersection}} \\ \cline{3-12} 
\multicolumn{2}{l}{} & Acc. & \multicolumn{1}{c}{Macro F1} & Acc. & \multicolumn{1}{c}{Macro F1} & Acc. & \multicolumn{1}{c}{Macro F1} & Acc. & \multicolumn{1}{c}{Macro F1} & Acc. & \multicolumn{1}{c}{Macro F1} \\ \hline
\multicolumn{2}{l}{Logistic Regression} & \multicolumn{1}{r}{0.707} & 0.686 & \multicolumn{1}{r}{0.684} & 0.673 & \multicolumn{1}{r}{0.721} & 0.675 & \multicolumn{1}{r}{0.737} & \textbf{0.686} & \multicolumn{1}{r}{0.800} & 0.787 \\ \hline
\multicolumn{2}{l}{Decision Tree} & \multicolumn{1}{r}{0.657} & 0.625 & \multicolumn{1}{r}{0.644} & 0.637 & \multicolumn{1}{r}{0.667} & 0.611 & \multicolumn{1}{r}{0.703} & 0.627 & \multicolumn{1}{r}{0.733} & 0.719 \\ \hline
\multicolumn{2}{l}{Random Forest} & \multicolumn{1}{r}{\textbf{0.767}} & \textbf{0.724} & \multicolumn{1}{r}{\textbf{0.717}} & \textbf{0.707} & \multicolumn{1}{r}{\textbf{0.814}} & \textbf{0.738} & \multicolumn{1}{r}{\textbf{0.744}} & 0.549 & \multicolumn{1}{r}{\textbf{0.815}} & \textbf{0.799} \\ \hline \hline
\multicolumn{2}{l}{\texttt{RoBERTa-base}} & \multicolumn{1}{r}{0.816} & 0.771 & \multicolumn{1}{r}{0.756} & 0.738 & \multicolumn{1}{r}{0.838} & 0.743 & \multicolumn{1}{r}{0.701} & 0.415 & \multicolumn{1}{r}{0.811} & 0.781 \\ \hline
\end{tabular}%
}
\caption{Class-wise distribution and classification results for different LLMs. "Intersection" refers to questions always or never solved correctly by all or any LLM, respectively. All classification results are mean values across 5 runs with varying initialization seeds. The best results for feature-based classifiers are highlighted in bold.}
\label{tab:class_distr_and_classification}
\end{table*}

\subsection{Success Rate Distribution}

\begin{figure}[t!]
     \centering
     \begin{subfigure}[b]{0.5\textwidth}
         \centering
         \includegraphics[width=\textwidth]{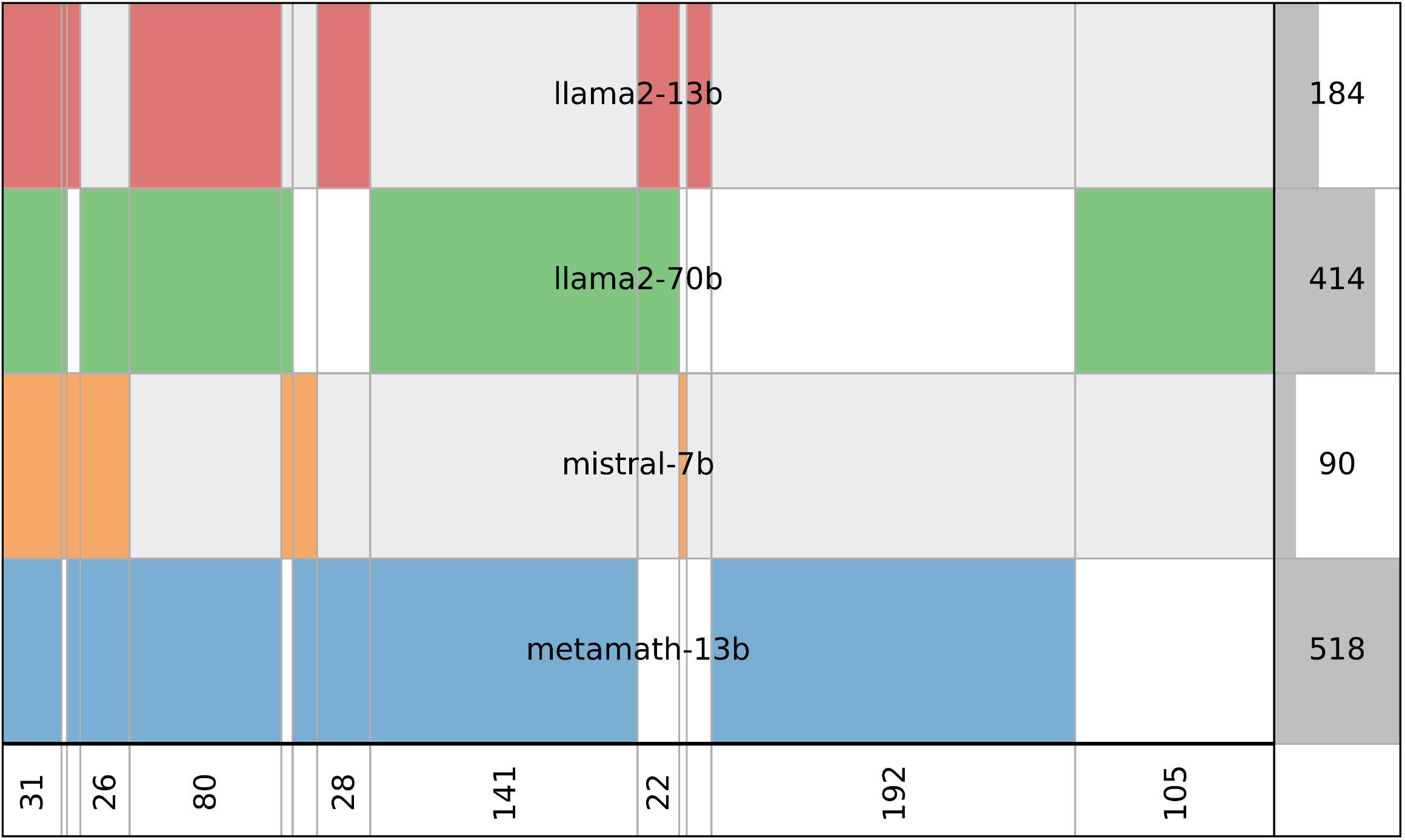}
         \caption{Always correct}
         \label{fig:sr_venn_a}
     \end{subfigure} %
     \begin{subfigure}[b]{0.5\textwidth}
         \centering
         \includegraphics[width=\textwidth]{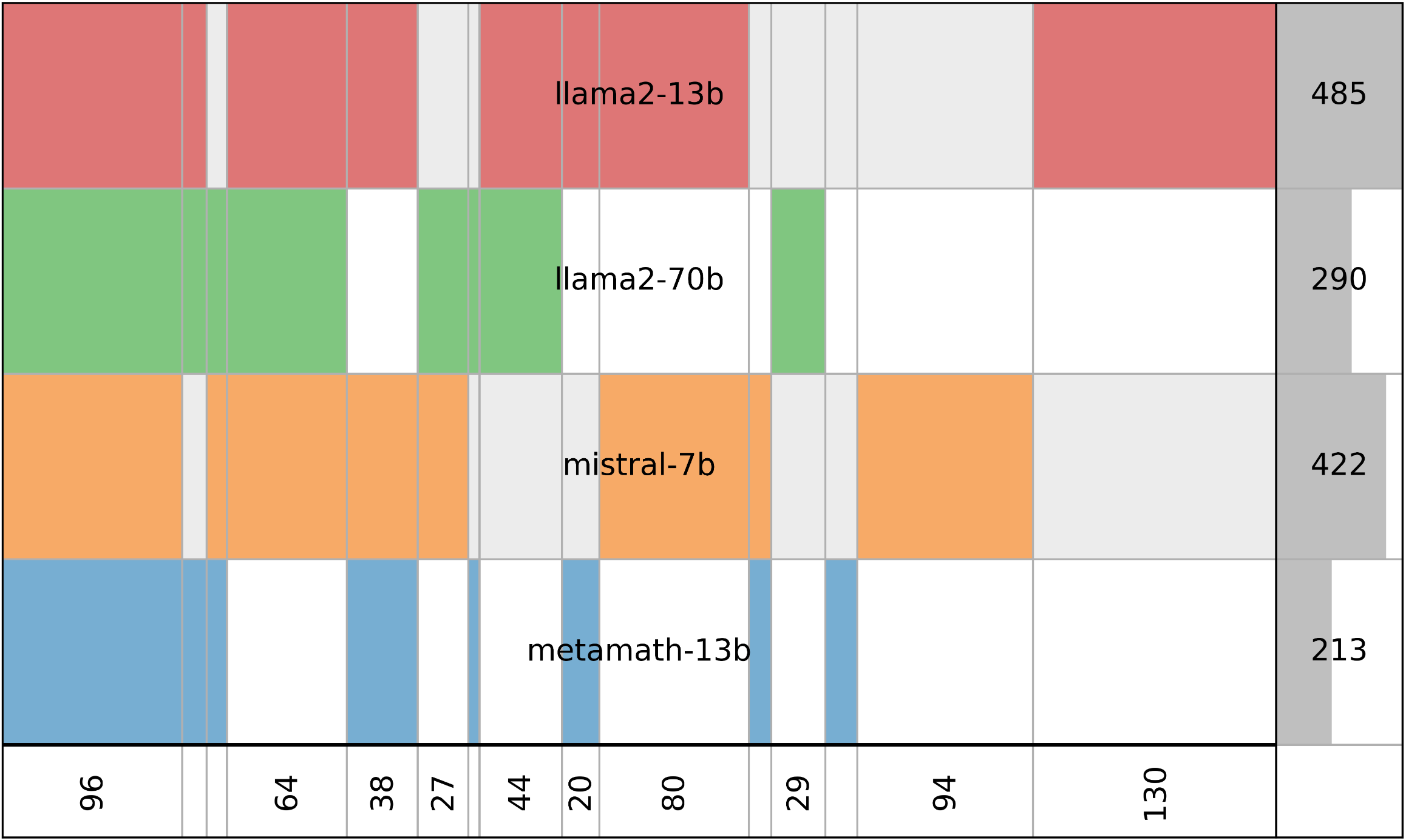}
         \caption{Never correct}
         \label{fig:sr_venn_b}
     \end{subfigure}
        \caption{Number of questions from the {\tt GSM8K}-test (a) always and (b) never answered correctly by each LLM. The rows in each figure correspond to individual LLMs, with the counts on the right denoting the total number of questions always (or never) answered correctly by each LLM. The counts at the bottom denote the number of questions in each subset of LLMs.} 
        \label{fig:sr_venn}
\end{figure}

We report the mean success rates for each LLM on {\tt GSM8K}'s test set in Table \ref{tab:success_rates}.\footnote{Our results generally align with those reported previously for these models.
} We observe that {\tt Llama2} 13B and 70B follow the expected order of scores along their respective parameter counts. {\tt Mistral-7B} scores similar to the 13B {\tt Llama2} model, and the additional fine-tuning allows {\tt MetaMath-13B} to outperform the other models (including the 70B {\tt Llama2}). Figures \ref{fig:sr_venn_a} and \ref{fig:sr_venn_b} respectively capture the number of questions \textit{always} and \textit{never} answered correctly by each LLM. Overall, {\tt MetaMath-13B} has the lowest number of incorrectly and the highest number of correctly answered questions across the tested LLMs.

\subsection{Classification Results}

To compare classifiers' performance, we report the accuracy and macro-F1 scores for each classifier and LLM-specific test data split (see Table \ref{tab:class_distr_and_classification}). We observe that Random Forest outperforms other classifiers across most solution sets. 

At the same time, we also note that, due to significant class imbalance, this task is not easy for the classifiers, with the best accuracy scores across LLM splits being in the range of $71.7\%-81.4\%$. The small number of questions always or never solved correctly by any LLM speaks to the models' varying capabilities (and potential points of brittleness). We include additional analysis of the results in Appendix \ref{sec:appendix_results_analysis}.

For comparison, we also report the classification results for a fine-tuned RoBERTa-base model~\cite{liu2019roberta} for the same training and evaluation sets (tuned on the question and gold solution as input text; see Appendix \ref{sec:appendix_training_details} for more details) in Table \ref{tab:class_distr_and_classification}. We note that the Transformer base classifier scores on a par or a few points above the best statistical classifier, i.e., Random Forest, suggesting that the proposed feature-based classifiers are not far behind token-level contextual models for this task.

\subsubsection{Feature Importance}
\label{subsec:feature_importance}

The statistical classifiers used in our experiments allow us to estimate the importance of each feature and its contribution to the classification performance. We report the top $10$ features with the highest aggregate ranks across LLM data splits and classifiers in Table \ref{tab:feature_importance}. We use mean rank here as a proxy for relative importance across features, and the respective standard deviations indicate how spread out this importance is across classifiers and queried LLMs. We observe that a greater number (\texttt{Gx\_op\_unique\_count}) and diversity (\texttt{Gx\_op\_diversity}) in math operations, and the use of infrequent numerical tokens in the question and solution body (\texttt{Qx\_} \& \texttt{Gx\_mean\_numerical\_word\_rank}) impact the success rate. The list also contains linguistic features based on the phrasing of the questions: longer questions with a high number of noun phrases (\texttt{Qx\_np\_count}), mean syntactic depth (\texttt{Qx\_constituency\_tree\_depth}), and readability grade (\texttt{Qx\_flesch\_kinkaid\_grade}) are also difficult for LLMs to solve. Additionally, the need for extraneous information (\texttt{Gx\_world\_knowledge}), such as conversion units for time, distance, or weight, can make a question challenging. We also report value thresholds at which each feature affects the success rate significantly: see the results of the Student's {\em t}-test and {\em p}-values in Table \ref{tab:t_test} in Appendix \ref{sec:appendix_results_analysis}.

\begin{table}[]
\centering
\resizebox{0.85\columnwidth}{!}{%
\begin{tabular}{@{}llcc@{}}
\toprule
\multirow{2}{*}{\textbf{Type}} & \multicolumn{1}{c}{\multirow{2}{*}{\textbf{Feature Name}}} & \multicolumn{2}{c}{\textbf{Rank} ($N$=23)} \\ \cmidrule(l){3-4} 
 & \multicolumn{1}{c}{} & $\mu$ & $\sigma$ \\ \midrule
L & Qx\_np\_count & 1.2 & 0.45 \\ \midrule
M & Qx\_mean\_numerical\_word\_rank & 4 & 1.87 \\ \midrule
M & Gx\_op\_unique\_count & 4 & 2.65 \\ \midrule
M & Gx\_op\_diversity & 4.4 & 2.30 \\ \midrule
M & Gx\_mean\_numerical\_word\_rank & 4.4 & 1.82 \\ \midrule
L & Qx\_mean\_word\_rank & 5.6 & 1.82 \\ \midrule
L & Qx\_flesch\_kinkaid\_grade & 6 & 1.87 \\ \midrule
W & Gx\_world\_knowledge & 7.8 & 2.28 \\ \midrule
L & Qx\_constituency\_tree\_depth & 9.6 & 1.95 \\ \midrule
M & Gx\_op'+'\_count & 11.6 & 3.97 \\ \bottomrule
\end{tabular}%
}
\caption{Feature importance ranks across classification models and LLM-wise data subsets.}
\label{tab:feature_importance}
\end{table}

\subsubsection{Ablation Studies}

\begin{figure}[t]
     \centering
     \includegraphics[width=0.9\columnwidth]{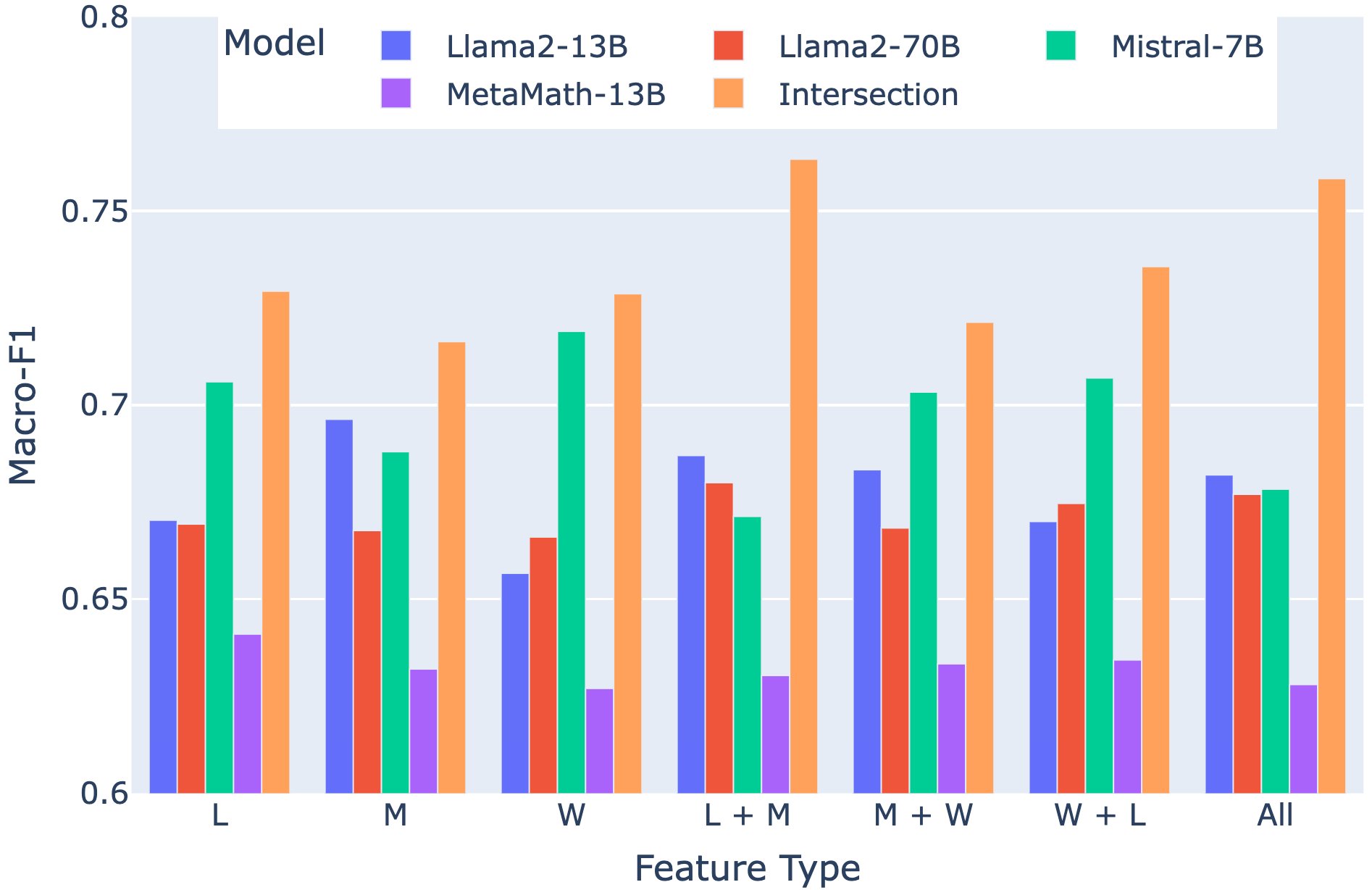}
     \caption{Results of the ablation studies across feature types ({\tt L} -- Linguistic, {\tt M} -- Mathematical, {\tt W} -- World Knowledge \& NLU). Each bar represents the mean macro-F1 score over all three classifier models.}
     \label{fig:ablation}
\end{figure}

To further measure the impact of each feature type, we report classification scores along different feature-type subsets in Figure \ref{fig:ablation}. We note that the feature set with all types (\texttt{L+M+W}) is not optimal for classification. For instance, the questions answered by {\tt Llama2-13B} are best classified using only mathematical features ({\tt M}). The best-performing classifiers for {\tt Llama2-7B,} {\tt MetaMath-13B}, and the intersection set either solely use linguistic features (\texttt{L}) or both linguistic and math features (\texttt{L+M}), whereas the world knowledge \& NLU feature set if sufficient for {\tt Mistral-7B}.   

\subsubsection{Impact of Linguistic Features}
In order to better gauge the impact of linguistic features on the success rate, we cluster questions by mathematical features. We fit a KMeans clustering model\footnote{\url{https://scikit-learn.org/stable/modules/generated/sklearn.cluster.KMeans.html}} on all math features for each question in the {\tt GSM8K} training set with a target cluster count of $100$. This helps group together questions from the data, wherein the math features hardly vary within each question subset (or cluster). Thus, variations in success rate across the questions within a cluster can be more clearly attributed to other, i.e., linguistic types of features. We report some notable Spearman correlation values between the linguistic feature values within a cluster and the corresponding success rates in Table \ref{tab:cluster_corr}. The strong and significant feature-wise negative correlations suggest that for a relatively fixed set of math features, questions with greater length, nesting, lexical rank, and reading grade become more challenging for LLMs to solve. Note that this form of analysis on feature-based minimal pairs is extractive in nature and may, to a certain extent, be restricted to the question types in the {\tt GSM8K} dataset. For a more exhaustive analysis for each feature, generative approaches to furnish question paraphrases with the desired set of linguistic features need to be employed.

\begin{table}[]
\resizebox{\columnwidth}{!}{%
\begin{tabular}{@{}llll@{}}
\toprule
\multicolumn{2}{l}{\textbf{Cluster}} & \multirow{2}{*}{\textbf{Feature}} & \multirow{2}{*}{\textbf{Spearman($\rho$)}} \\ \cmidrule(r){1-2}
ID & Size &  &  \\ \midrule
09 & 27 & Qx\_constituency\_tree\_depth & -0.64*** \\ \midrule
24 & 14 & Qx\_mean\_word\_rank & -0.82*** \\ \midrule
\multirow{2}{*}{63} & \multirow{2}{*}{62} & Qx\_token\_length & -0.42*** \\ \cmidrule(l){3-4} 
 &  & Qx\_word\_length & -0.43*** \\ \midrule
96 & 51 & Qx\_flesch\_kinkaid\_grade & -0.51*** \\ \bottomrule
\end{tabular}%
}
\caption{Cluster-wise feature correlations. The cluster count represents the number of questions included in the respective cluster. The p-value for all reported correlation values is \textless{}0.001 (marked by `***').}
\label{tab:cluster_corr}
\end{table}

\section{Conclusions}
\begin{figure}[h!]
    \centering
    \includegraphics[width=1.0\columnwidth]{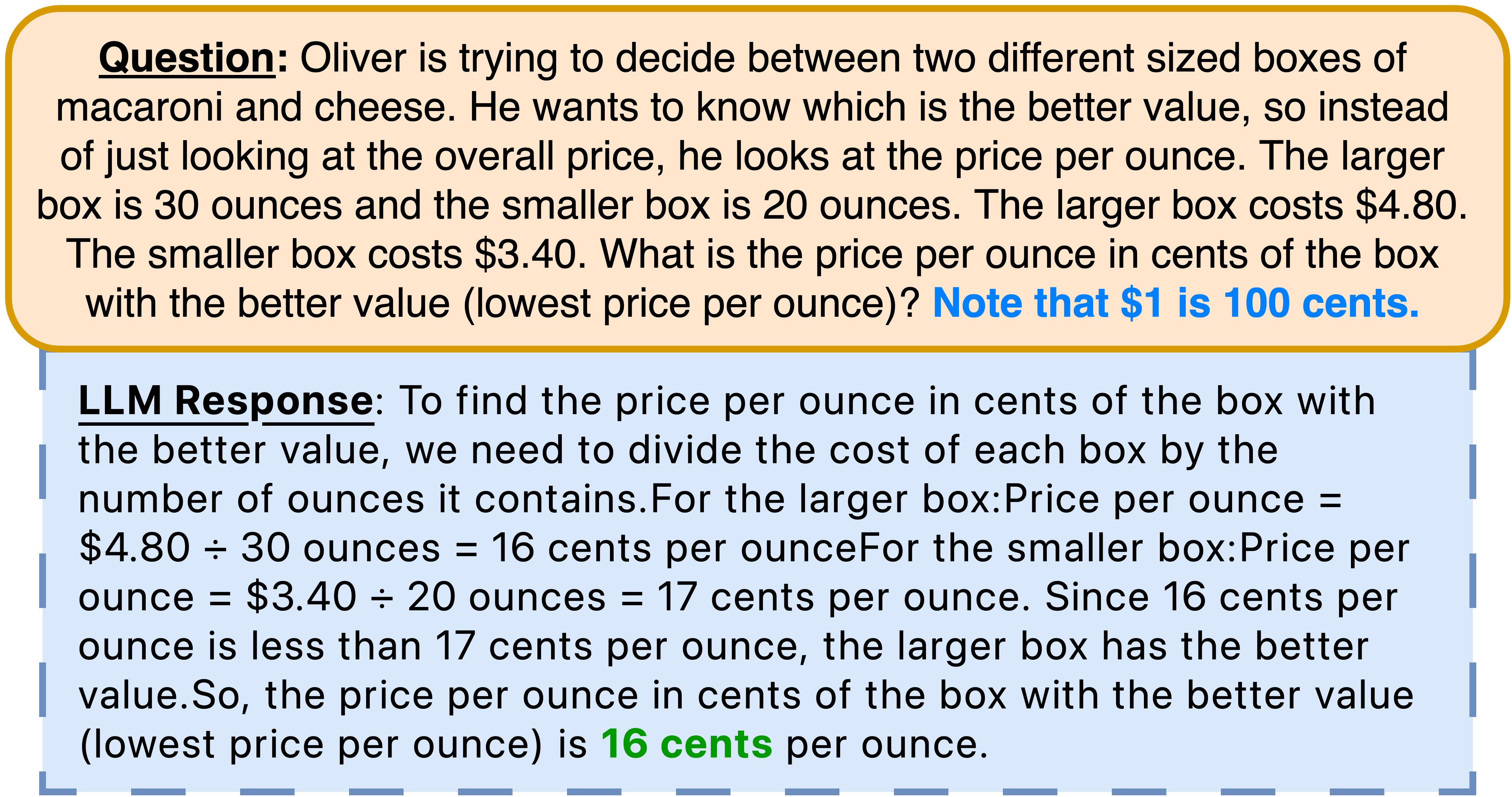}
     \caption{Solution attempt by {\tt Llama2-70B} on the question from Figure \ref{fig:sample_question}, with the required real-world knowledge explicitly specified.} 
     \label{fig:revised_question}
\end{figure}

This work aims to identify what aspects of MWPs make them difficult for LLMs to solve. To this end, we extract key features (spanning linguistic, mathematical, and real-world knowledge \& NLU-based aspects) to predict whether several LLMs can reliably solve MWPs from {\tt GSM8K}. We find that questions with a high number and diversity of math operations using infrequent numerical tokens are particularly challenging to solve. In addition, we show that lengthy questions with low readability scores and those requiring real-world knowledge are also seldom solved correctly. Our future work will rely on these findings to make informed modifications to questions in order to study the impact on LLMs' reasoning and MWP-solving abilities. Figure \ref{fig:revised_question} provides an example of an informed modification, which leads to improved LLM performance.

\section*{Limitations}

With the rapidly growing body of research on LLMs, this study necessarily has several limitations, which we discuss below.

\paragraph{Limited set of LLMs tested} We consider it important to test and report results with a diverse set of open-source LLMs, which motivated the selection of the specific models included in this study. At the same time, we do not claim this study to be comprehensive with respect to the range of LLMs tested and in future work, we plan to include more LLMs in this research.

\paragraph{Limited number of classification models} As the main goal of this study is to identify aspects of the MWPs that make them challenging for LLMs to solve, we have opted for a feature-based approach and a range of traditional classification models as opposed to less transparent but more powerful black-box algorithms. Our results show that the prediction task is challenging for the traditional classifiers that we used, and it is likely that these results can be improved with stronger classification models.

\paragraph{Limitations of the dataset} In this work, we have focused on a single MWP dataset ({\tt GSM8K}) due to its unique properties, namely the high quality of the questions, high diversity of the tasks (including linguistic diversity of the questions), and moderate difficulty of the math problems covered~\cite{cobbe2021training}. At the same time, we recognize that the results we report in this work may be limited in certain ways to the dataset on which we report them. Our future work will apply this approach to other available MWP datasets~\cite{kim-etal-2023-aint,wang2023msat,patel-etal-2021-nlp,miao-etal-2020-diverse} to verify the consistency of the findings.

\paragraph{Impact studies} Finally, whilst we have identified aspects of the MWPs that make them challenging for LLMs to solve, we admittedly presented only one example (see Figure \ref{fig:revised_question}) where acting upon one of the identified aspects improves the output of an LLM. While a thorough investigation of the impact of such informed modifications is outside the scope of the current paper, such experiments will follow in future work to demonstrate the practical usefulness of the identified MWP aspects.

\section*{Ethics Statement}

We foresee no serious ethical implications from this study.

\section*{Acknowledgments}
We are grateful to the Campus Super Computing Center at MBZUAI for supporting this work. We also thank the anonymous reviewers for their valuable feedback.

\bibliography{anthology,custom}

\appendix

\newpage

\section{Feature Details}
\label{sec:appendix_feature_details}

Below, we describe the features used in our study and how they were extracted.

\begin{table*}[ht!]
\centering
\resizebox{0.7\textwidth}{!}{%
\begin{tabular}{@{}ccclcrr@{}}
\toprule
\textbf{Type} & \textbf{Source} & \textbf{\#} & \multicolumn{1}{c}{\textbf{Feature Name}} & \textbf{Range} & \textbf{$\mu$} & \textbf{$\sigma$} \\ \midrule
\multirow{9}{*}{L} & Q & 1 & Qx\_token\_length & {[}12 -- 239{]} & 66.05 & 24.384 \\ \cmidrule(l){2-7} 
 & Q & 2 & Qx\_sentence\_length & {[}1 -- 13{]} & 3.431 & 1.201 \\ \cmidrule(l){2-7} 
 & Q & 3 & Qx\_word\_length & {[}9 -- 184{]} & 45.885 & 17.832 \\ \cmidrule(l){2-7} 
 & Q & 4 & Qx\_flesch\_kinkaid\_grade & {[}-1.9 -- 26.3{]} & 4.236 & 2.468 \\ \cmidrule(l){2-7} 
 & Q & 5 & Qx\_mean\_word\_rank & {[}3661.96 -- 21929.96{]} & 10646.615 & 2110.891 \\ \cmidrule(l){2-7} 
 & Q & 6 & Qx\_constituency\_tree\_depth & {[}5 -- 31{]} & 10.803 & 2.798 \\ \cmidrule(l){2-7} 
 & Q & 7 & Qx\_np\_count & {[}3 -- 74{]} & 18.034 & 7.488 \\ \cmidrule(l){2-7} 
 & Q & 8 & Qx\_prp\_count & {[}0 -- 16{]} & 1.772 & 1.854 \\ \cmidrule(l){2-7} 
 & Q & 9 & Qx\_coref\_count & {[}0 -- 16{]} & 0.462 & 1.283 \\ \midrule
\multirow{13}{*}{M} & Q & 10 & Qx\_arg\_count & {[}0 -- 17{]} & 4.438 & 1.94 \\ \cmidrule(l){2-7} 
 & Q & 11 & Qx\_word\_arg\_count & {[}0 -- 14{]} & 1.091 & 1.397 \\ \cmidrule(l){2-7} 
 & Q & 12 & Qx\_mean\_numerical\_word\_rank & {[}259.0 -- 29905.38{]} & 22643.319 & 3260.09 \\ \cmidrule(l){2-7} 
 & G & 13 & Gx\_arg\_count & {[}6 -- 73{]} & 24.377 & 9.732 \\ \cmidrule(l){2-7} 
 & G & 14 & Gx\_op`+'\_count & {[}0 -- 12{]} & 1.06 & 1.212 \\ \cmidrule(l){2-7} 
 & G & 15 & Gx\_op`-'\_count & {[}0 -- 6{]} & 0.601 & 0.78 \\ \cmidrule(l){2-7} 
 & G & 16 & Gx\_op`*'\_count & {[}0 -- 8{]} & 1.369 & 1.183 \\ \cmidrule(l){2-7} 
 & G & 17 & Gx\_op`/'\_count & {[}0 -- 7{]} & 0.621 & 0.789 \\ \cmidrule(l){2-7} 
 & G & 18 & Gx\_op`('\_count & {[}0 -- 4{]} & 0.026 & 0.187 \\ \cmidrule(l){2-7} 
 & G & 19 & Gx\_op\_unique\_count & {[}0 -- 6{]} & 2.284 & 0.93 \\ \cmidrule(l){2-7} 
 & G & 20 & Gx\_op\_diversity & {[}0.15 -- 1.0{]} & 0.758 & 0.196 \\ \cmidrule(l){2-7} 
 & G & 21 & Gx\_mean\_numerical\_word\_rank & {[}22645.0 -- 29915.0{]} & 28626.04 & 776.73 \\ \cmidrule(l){2-7} 
 & B & 22 & Gx\_parameter\_usage & {[}0.07 -- 1.0{]} & 0.642 & 0.241 \\ \midrule
W & B & 23 & Gx\_world\_knowledge & {[}0 -- 8{]} & 1.104 & 1.006 \\ \bottomrule
\end{tabular}%
}
\caption{Details of formulation and distribution (across {\tt GSM8K}) for all features included in the feature set. Each feature is of type: Linguistic (L), Mathematical (M), or World Knowledge and NLU (W) and is sourced either from the question body (Q), gold solution body (G), or both (B).}
\label{tab:feature_set}
\end{table*}

\noindent {\bf (1) Linguistic features (L)} include $9$ features pertaining to the question (Q) itself:
\begin{itemize}
    \item {\bf Qx\_token\_length}: The number of tokens in the tokenized version of the question body. We apply each LLM's respective tokenizer from \href{https://huggingface.co/}{{\tt HuggingFace}} to extract this feature. 
    \item {\bf Qx\_sentence\_length}: The number of sentences detected in the question body. We use the \href{https://pypi.org/project/sentence-splitter/}{\texttt{sentence\_splitter}} Python library to extract this count.
    \item {\bf Qx\_word\_length}: The number of space-separated segments (words) in the question body.
    \item {\bf Qx\_flesch\_kinkaid\_grade}: The readability grade of the question body as per the FKGL metric~\cite{flesch1948new}. We use the \href{https://pypi.org/project/textstat/}{{\tt textstat}} Python library to extract this feature.
    \item {\bf Qx\_mean\_word\_rank}: The mean vocabulary rank (in decreasing order of frequency) of the tokens in the question body. We use the same tokenizer set used for \texttt{Qx\_token\_length}.
    \item {\bf Qx\_constituency\_tree\_depth}: The mean depth of the constituency tree across the sentences in the question body. We use Stanford's \href{https://stanfordnlp.github.io/stanza/}{{\tt Stanza}} parsing library to parse the questions.
    \item {\bf Qx\_np\_count}: Number of distinct noun phrases detected in the question body. We extract this from the constituency parse collected from the {\tt Stanza} parser. 
    \item {\bf Qx\_prp\_count}: Number of prepositions in the question body. We use the part-of-speech tags generated as part of the parse by {\tt Stanza}.
    \item {\bf Qx\_coref\_count}: Number of pronominal or nominal instances of coreference in the question body. We use Stanford's \href{https://stanfordnlp.github.io/CoreNLP/coref.html}{{\tt CorefAnnotator}} to extract this feature.
\end{itemize}

\noindent {\bf (2) Mathematical features (M)} include $12$ features pertaining to the question (Q) and gold solution (G):
\begin{itemize}
    \item {\bf Qx\_arg\_count}: The number of distinct numerical quantities (e.g., ``\textbf{3.5} hours later'' or ``\textbf{100} boxes'') in the question body. We use a {\tt Regexp} pattern to detect whole numbers, decimal point values, and quantities preceded by a negative sign or dollar (and other currency) signs.
    \item {\bf Qx\_word\_arg\_count}: The number of quantities mentioned in word-form (``\textbf{three} times'' or ``\textbf{half} as much'') in the question body. We use a vocabulary of frequently used word-form tokens and accommodate compound expressions (e.g., ``twenty-two'').
    \item {\bf Qx\_mean\_numerical\_word\_rank}: The mean vocabulary rank of the numerical tokens in the question body. We first isolate numerical tokens tokenized by respective tokenizers, then aggregate their token rank.
    \item {\bf Gx\_arg\_count}: The number of distinct numerical quantities present as plain text or on the left-hand side of equations in the gold solution. We use the same {\tt Regexp} pattern used for \texttt{Qx\_arg\_count}.
    \item {\bf Gx\_op\{`+'/ `-'/ `*'/ `/'/ `('\}\_count}: Number of times each listed math operation is used in the gold solution. A simple {\tt Regexp} pattern is applied to extract these from within equations.
    \item {\bf Gx\_op\_unique\_count}: The maximum number of times a single operation has been used in the gold solution. For instance, ``$3+4.5+7+1-2.7$'' contains 3 instances of the `+' operator.
    \item {\bf Gx\_op\_diversity}: Ratio of the number of unique math operators used to the total number of operators in the gold solution. For instance, a question with the consolidated math solution expression "$(2\times12)\times3=72$" contains two arithmetic operations in total but only one unique operation type, i.e., `$\times$,' \texttt{Gx\_op\_diversity}$=1/2=0.5$. 
    \item {\bf Gx\_mean\_numerical\_word\_rank}: The mean vocabulary rank of the numerical tokens used on the left-hand side of equations in the gold solution. Extracted the same way as \texttt{Qx\_mean\_numerical\_word\_rank}.
    \item {\bf Bx\_parameter\_usage}: The ratio of distinct arguments used in the gold solution to that in the question body. A value lower than 1 indicates that one or more arguments provided in the question were not required to solve the MWP (potentially acting as distractors).
\end{itemize}

\noindent {\bf (3) World knowledge and NLU features (W)} include:
\begin{itemize}
    \item {\bf Bx\_world\_knowledge}: The number of distinct arguments on the left-hand-side of equations in the gold solution, that are neither present in the question body nor produced as intermediate results from any prior equations in the solution. A non-zero value is interpreted as the use of a quantity (perhaps a conversion factor, or the number of entities involved in computing mean) unspecified by the question. The arguments were extracted from both sides using the same {\tt Regexp} policies used for previous features.
\end{itemize}

Table \ref{tab:feature_set} shows further statistics on the features, including the  range as well as the mean and standard deviation of the values for each feature type. Additionally, we report the Spearman correlation between all pairs in the feature set in Figure \ref{fig:feature_correlation}.

\begin{figure*}[h!]
    \centering
    \includegraphics[width=1.0\textwidth]{fig_feature_correlation_v2.png}
     \caption{Spearman correlation matrix between features. All correlation values are marked with `*', `**', and `***' if their corresponding p-values are less than $0.05$, $0.01$, and $0.001$ respectively.} 
     \label{fig:feature_correlation}
\end{figure*}

\newpage
\section{Querying Details}
\label{sec:appendix_querying_details}

\subsection{Prompt Template}
\begin{figure}[h!]
    \centering
    \includegraphics[width=1.0\columnwidth]{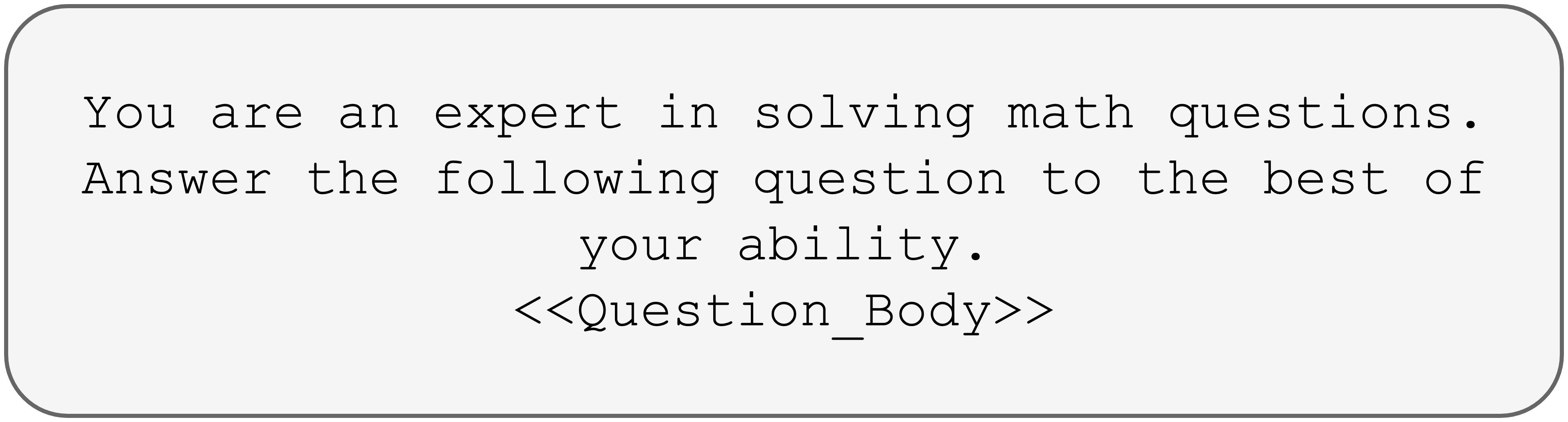}
     \caption{Prompt template used for solution generation across LLMs.} 
     \label{fig:prompt_template}
\end{figure}

We use a simple task-specific prompt (see Figure \ref{fig:prompt_template}) either prepended to the question-prompt or specified as a system-prompt if an LLM input query format requires so. 

\subsection{LLM Details}
The exact large language models used in our experiments, along with their reported performance on {\tt GSM8K} according to the OpenLLM leaderboard ~\cite{open-llm-leaderboard} are mentioned in Table \ref{tab:llm_background}. All LLMs and libraries used are open-source. The license to use Meta's {\tt Llama2} models was procured through due process.

\begin{table}[h!]
\resizebox{\columnwidth}{!}{%
\begin{tabular}{lcc}
\hline
\textbf{Model} & \textbf{HuggingFace Model Name} & \textbf{Pass@1} \\ \hline
Llama2-13B & \texttt{meta-llama/Llama-2-13b-chat-hf} & 28.70 \\ \hline
Llama2-70B & \texttt{meta-llama/Llama-2-70b-chat-hf} & 56.80 \\ \hline
Mistral-7B & \texttt{mistralai/Mistral-7B-Instruct-v0.2} & 40.03 \\ \hline
MetaMath-13B & \texttt{meta-math/MetaMath-13B-V1.0} & 72.30 \\ \hline
\end{tabular}%
}
\caption{List of HuggingFace model variants and their respective reported pass@1 (single run) accuracies on the {\tt GSM8K} test set from the OpenLLM leaderboard.}
\label{tab:llm_background}
\end{table}

\subsection{Implementation and Compute Resources Used}
We use \href{https://github.com/vllm-project/vllm}{vLLM} to load and query models. Models of parameter sizes 7B and 13B were queried with a single NVIDIA A100 GPU. {\tt Llama2-70B} was loaded and queried using 4 A100 GPUs. Each query was set to a temperature of $0.8$, and a maximum token length of $2000$. Each question was queried $5$ times by each LLM, with a varying seed. Querying the entire {\tt GSM8K} dataset ($8,793$ questions) took approximately 1 hour for each LLM.

\begin{table*}[!h]
\resizebox{\textwidth}{!}{%
\begin{tabular}{lccrccrccrccrccr}
\hline
\multirow{2}{*}{\textbf{Feature}} & \multicolumn{3}{c}{\textbf{Llama2-13B}} & \multicolumn{3}{c}{\textbf{Llama2-70B}} & \multicolumn{3}{c}{\textbf{Mistral-7B}} & \multicolumn{3}{c}{\textbf{MetaMath-13B}} & \multicolumn{3}{c}{\textbf{Intersection}} \\ \cline{2-16} 
 & Thresh. & Diff. & \multicolumn{1}{c}{T-val} & Thresh. & Diff. & \multicolumn{1}{c}{T-val} & Thresh. & Diff. & \multicolumn{1}{c}{T-val} & Thresh. & Diff. & \multicolumn{1}{c}{T-val} & Thresh. & Diff. & \multicolumn{1}{c}{T-val} \\ \hline
Gx\_num\_arg\_count & 51.102 & 0.313 & -8.328 & 53.273 & 0.380 & -7.747 & 55.082 & 0.283 & -5.452 & 59.061 & 0.438 & -5.558 & 57.735 & 0.330 & -5.977 \\ \hline
Qx\_np\_count & 40.673 & 0.200 & -5.440 & 47.465 & 0.435 & -5.090 & 33.429 & 0.207 & -11.201 & 47.918 & 0.448 & -7.230 & 47.918 & 0.333 & -5.645 \\ \hline
Gx\_arg\_count & 57.959 & 0.336 & -6.018 & 58.111 & 0.380 & -5.742 & 57.959 & 0.289 & -5.644 & 51.122 & 0.271 & -10.323 & 60.694 & 0.335 & -5.249 \\ \hline
Qx\_word\_length & 91.143 & 0.219 & -7.944 & 97.384 & 0.300 & -8.468 & 19.714 & 0.233 & -7.016 & 101.857 & 0.275 & -9.267 & 116.143 & 0.251 & -5.328 \\ \hline
Gx\_op'-'\_count & 2.082 & 0.198 & -7.683 & 2.061 & 0.246 & -9.009 & 2.082 & 0.181 & -7.660 & 3.061 & 0.339 & -7.348 & 3.061 & 0.280 & -6.383 \\ \hline
Qx\_prp\_count & 7.184 & 0.215 & -5.957 & 8.081 & 0.313 & -6.322 & 8.163 & 0.235 & -5.475 & 7.184 & 0.167 & -6.039 & 8.163 & 0.233 & -6.834 \\ \hline
Qx\_sentence\_length & 6.143 & 0.206 & -6.843 & 6.091 & 0.264 & -8.278 & 6.143 & 0.203 & -7.357 & 7.122 & 0.253 & -5.756 & 7.122 & 0.229 & -5.473 \\ \hline
Gx\_op'+'\_count & 4.163 & 0.241 & -7.849 & 4.121 & 0.293 & -8.996 & 5.143 & 0.296 & -6.038 & 3.184 & 0.080 & -5.377 & 5.143 & 0.238 & -6.081 \\ \hline
Qx\_token\_length & 123.184 & 0.211 & -8.937 & 126.646 & 0.258 & -9.557 & 113.918 & 0.214 & -12.560 & 132.449 & 0.232 & -9.953 & 118.551 & 0.219 & -14.125 \\ \hline
Gx\_op'*'\_count & 4.082 & 0.211 & -6.135 & 4.040 & 0.220 & -6.043 & 4.082 & 0.184 & -5.831 & 5.061 & 0.268 & -5.423 & 5.061 & 0.250 & -5.295 \\ \hline
\end{tabular}%
}
\caption{Feature-wise thresholds which reflect the greatest difference in the corresponding mean success rate. For each feature, the optimal threshold creates two sets of questions on either side, wherein the difference in the corresponding mean success rates of the two sets is the greatest. We perform Student's $t$-tests on both sets to determine if this difference is significant and report the corresponding $t$ values. All results reported in the table have an absolute $t$-value \textgreater 5 and a $p$-value \textless 0.0001.}
\label{tab:t_test}
\end{table*}

\begin{table*}[]
\centering
\resizebox{\textwidth}{!}{%
\begin{tabular}{lcccccccccc}
\hline
\multicolumn{1}{c}{\multirow{2}{*}{\textbf{Classification Model}}} & \multicolumn{2}{c}{\textbf{Llama2-13B}} & \multicolumn{2}{c}{\textbf{Llama2-70B}} & \multicolumn{2}{c}{\textbf{Mistral-7B}} & \multicolumn{2}{c}{\textbf{MetaMath-13B}} & \multicolumn{2}{c}{\textbf{Intersection}} \\ \cline{2-11} 
\multicolumn{1}{c}{} & F1(Never) & F1(Always) & F1(Never) & F1(Always) & F1(Never) & F1(Always) & F1(Never) & F1(Always) & F1(Never) & F1(Always) \\ \hline
Logistic Regression & 0.8454 & 0.4968 & 0.5641 & \textbf{0.7862} & 0.8926 & 0.5081 & 0.0095 & 0.8354 & 0.8263 & \textbf{0.7184} \\ \hline
Decision Tree & 0.8199 & 0.4532 & 0.4958 & 0.7431 & 0.8732 & 0.4592 & \textbf{0.0727} & \textbf{0.8373} & 0.732 & 0.6496 \\ \hline
Random Forest & \textbf{0.8535} & \textbf{0.5049} & \textbf{0.6133} & 0.7824 & \textbf{0.8936} & \textbf{0.5161} & 0.019 & 0.8361 & \textbf{0.8439} & 0.7216 \\ \hline
\end{tabular}%
}
\caption{Class-wise F1 scores for each classification model across LLM solution splits.}
\label{tab:class_wise_f1}
\end{table*}

\newpage
\section{Training Details}
\label{sec:appendix_training_details}

\subsection{Preprocessing}
Before training classifiers, we perform the following steps on the feature data:
\begin{enumerate}
    \item \textbf{Pruning}: For each training feature-set, we iteratively remove features with high correlation with other features until no two columns in the data have an absolute Spearman correlation higher than $0.5$.
    \item \textbf{Scaling}: We fit scikit-learn's \href{https://scikit-learn.org/stable/modules/generated/sklearn.preprocessing.StandardScaler.html}{{\tt StandardScaler}} onto the train split to normalize the mean and standard deviation of all features. We then apply the same scaler on the test split features.
    \item \textbf{Balancing}: As many LLM solution splits either contain too few always-correct or always-incorrect question samples, we use imblearn's \href{https://imbalanced-learn.org/stable/references/generated/imblearn.over_sampling.RandomOverSampler.html}{{\tt RandomOverSampling}} tool to balance the proportions of the two classes in each run.
\end{enumerate}

\subsection{Hyperparameter Search}
For each classifier model and LLM solution split pair, we conduct Bayesian optimization on the ranges of key hyperparameters for each classifier. As the objective function, we maximize the Macro-F1 score on a 15\% held-out set of \texttt{GSM8K}'s test split to prevent the models from over-fitting onto the train samples.

\subsection{Fine-tuning RoBERTa Classifier}
We use HuggingFace's \href{https://huggingface.co/docs/transformers/en/main_classes/trainer}{Trainer} module to tune a pre-trained \href{https://huggingface.co/FacebookAI/roberta-base}{RoBERTa-base} classifier on the same target data as the statistical classifiers in each setting. The corresponding input text for training and evaluation was built by concatenating the \texttt{GSM8K} question and gold-solution text for each sample, i.e., \texttt{"<question> {Question Body} </question> <solution> {Solution Body} </solution>"}. The model is trained for $3$ epochs with a peak learning rate of $2e-5$ and a warmup ratio of $0.1$. On a Tesla P100, each training run took approximately 10-15 minutes.

\newpage
\section{Results Analysis}
\label{sec:appendix_results_analysis}

\subsection{Class-wise Classification Review}
Table \ref{tab:class_wise_f1} reports the class-wise F1-scores (for questions that are always or never solved correctly) for each classifier across LLM question splits. Though there may be notable class imbalances among the two classes across splits, all classifiers were trained with proportional oversampling. We see that for relatively smaller pretrained models, i.e., {\tt Llama2-13B} and {\tt Mistral-7B}, F1 scores for always-incorrect questions are significantly higher than their counterparts. Thus, the scores indicate that for smaller models, questions answered incorrectly are more predictable. For larger models like {\tt Llama2-70B}, this difference is lower, with the always-correct questions being somewhat more predictable. For the fine-tuned {\tt MetaMath-13B} model, the small number of questions that are never answered correctly, fail to provide a generalizable sample for predicting on unseen data.

\subsection{Feature Impact}
We continue our discussion of feature importance (from Section \ref{subsec:feature_importance}) by identifying pivot points for key features about which the corresponding success rates for questions show a significant difference in mean values. We perform Student's $t$-tests on equally spaced thresholds along each feature and report the thresholds which show the highest variation in mean success rates in Table \ref{tab:t_test}. 

We see that across most LLMs, a significant rise in the mean success rate is observed as the question contains, on average, more than $6$-$7$ sentences, $90$-$115$ words, or $113$-$132$ tokens. We get a better idea of the kind of questions \texttt{Mistral-7B} gets wrong more often than other models, as its threshold ($19.71$ words) for the number of words in the question body is substantially lower than the average. Across features involving number of math operations, the threshold for the fine-tuned \texttt{MetaMath-13B} model is either on a par or higher than other models.

\end{document}